\documentclass[letterpaper]{article} 
\usepackage{aaai25}  
\usepackage{times}  
\usepackage{helvet}  
\usepackage{courier}  
\usepackage[hyphens]{url}  
\usepackage{graphicx} 
\usepackage{natbib}  
\usepackage{caption} 
\usepackage{algorithm}
\usepackage{algorithmic}
\usepackage{amsmath}
\usepackage{amssymb}
\usepackage{dsfont}
\usepackage{booktabs}
\usepackage{subfigure}
\usepackage{multirow} 
\usepackage[table]{xcolor}
\usepackage{soul}

\usepackage{newfloat}
\usepackage{listings}
\DeclareCaptionStyle{ruled}{labelfont=normalfont,labelsep=colon,strut=off} 
\floatstyle{ruled}
\newfloat{listing}{tb}{lst}{}
\floatname{listing}{Listing}
\title{D\&M: Enriching E-commerce Videos with Sound Effects by Key Moment Detection and SFX Matching}
\author{
    Jingyu Liu\textsuperscript{1,2}, 
    Minquan Wang\textsuperscript{2}, 
    Ye Ma\textsuperscript{2}, Bo Wang\textsuperscript{2}, 
    Aozhu Chen\textsuperscript{1,2}, 
    Quan Chen\textsuperscript{2}, 
    Peng Jiang\textsuperscript{2}, \\ Xirong Li\textsuperscript{1}\footnotemark[2]
}
\affiliations{
    \textsuperscript{1}Renmin University of China \\
    \textsuperscript{2}Kuaishou Technology \\
    liujingyu2023@ruc.edu.cn,~chenquan06@kuaishou.com
    
}

\usepackage{bibentry}

\usepackage{xcolor}

\newcommand{\etc}{\emph{etc.}~}

\newcommand{\ie}{\emph{i.e.}~}
\newcommand{\wrt}{\emph{w.r.t.}~}

\begin{document}

\maketitle

\renewcommand{\thefootnote}{\fnsymbol{footnote}}
\footnotetext[2]{Corresponding author (xirong@ruc.edu.cn)}
\footnotetext[3]{by Steven Spielberg}

\begin{abstract}
Videos showcasing specific products are increasingly important for E-commerce. Key moments naturally exist as the first appearance of a specific product, presentation of its distinctive features, the presence of a buying link, etc. Adding proper sound effects (SFX) to such moments, or \emph{video decoration with SFX} (VDSFX), is crucial for enhancing user engaging experience. Previous work adds SFX to videos by video-to-SFX matching at a holistic level, lacking the ability of adding SFX to a specific moment. Meanwhile, previous studies on video highlight detection or video moment retrieval consider only moment localization, leaving moment to SFX matching untouched. By contrast, we propose in this paper D\&M, a unified method that  accomplishes key moment detection and moment-to-SFX matching simultaneously. Moreover, for the new VDSFX task we build a large-scale dataset SFX-Moment from an E-commerce video creation platform. For a fair comparison, we build competitive baselines by extending a number of current video moment detection methods to the new task. Extensive experiments on SFX-Moment show the superior performance of the proposed method over the baselines. 
\end{abstract}

\begin{links}
\link{Code and Data}{https://rucmm.github.io/VDSFX/}
\end{links}

\section{Introduction} \label{sec:intro}
\begin{figure}[t!]
    \centering
    \subfigure[Input video] {\includegraphics[width=0.99\linewidth]{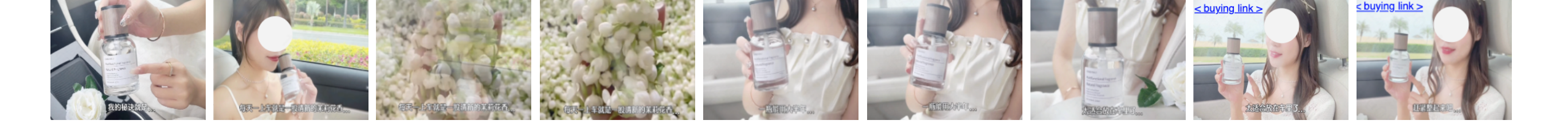}}
    \subfigure[Ground truth] {\includegraphics[width=0.99\linewidth]{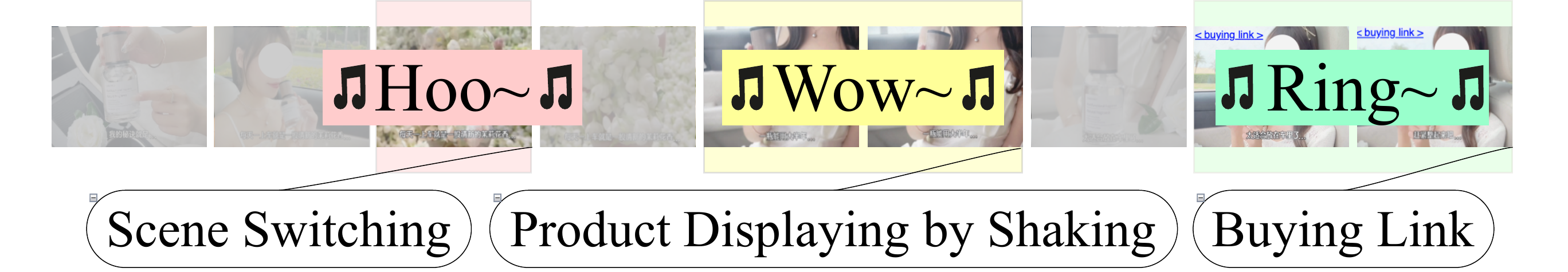}}
    \subfigure[Prediction by Moment-DETR+] {\includegraphics[width=0.99\linewidth]{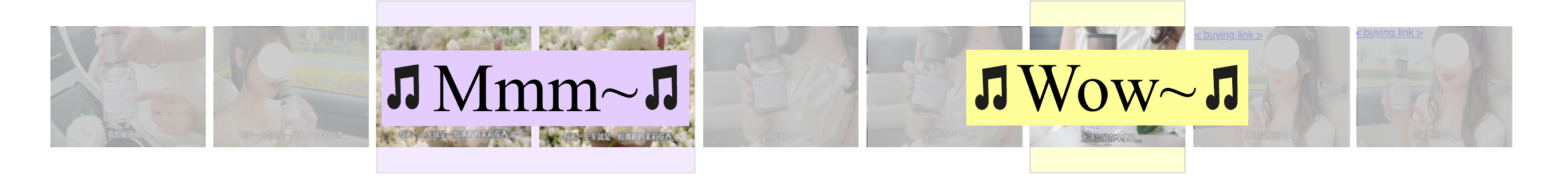}}
    \subfigure[Prediction by $\text{R}^2\text{-Tuning}$+] {\includegraphics[width=0.99\linewidth]{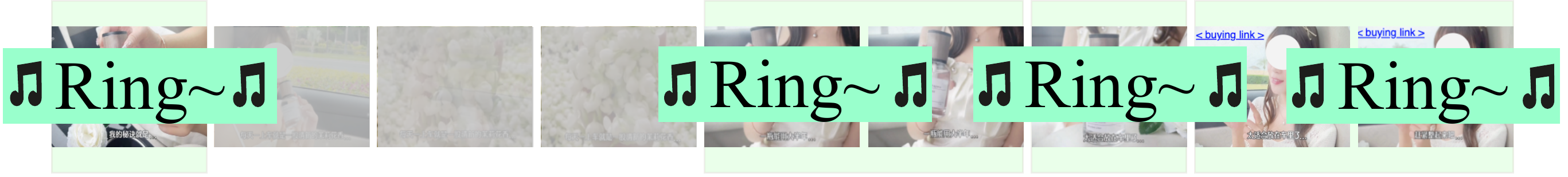}}
    \subfigure[Prediction by proposed D\&M] {\includegraphics[width=0.99\linewidth]{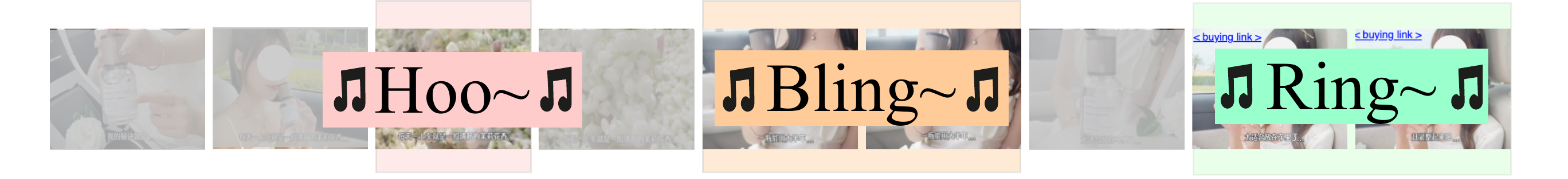}}
    \caption{Illustration of video decoration with sound effects (VDSFX), aiming to automatically add proper SFX to key moments, which are also auto-detected, in a given E-commerce video. 
    Moment-DETR+ and $\text{R}^2\text{-Tuning}$+ are baselines we implement, by re-purposing Moment-DETR \cite{lei2021detecting} and $\text{R}^2\text{-Tuning}$ \cite{liu2024r} for the new task, with their detected moments used for moment-to-SFX matching. Best viewed digitally.}
    \label{fig:intro}
\end{figure}

\textit{The eye sees better when the sound is great}\footnotemark[3]. In this paper we aim to automatically add sound effects (SFX) to key moments, which is also auto-detected, in E-commerce videos to enhance their expressiveness, see Fig. \ref{fig:intro}.


Videos showcasing specific products are increasingly important for E-commerce.  Key moments naturally exist as  the first appearance of a specific product, presentation of its major features, the presence of a buying link, \etc  Adding proper SFX to such moments is crucial for enhancing the engaging experience that helps turn a video viewer to a product consumer. Note that in contrast to a background music (BGM), a sound effect is a sound, \emph{not} music. The two differ from their purposes and durations. BGMs play subtly behind the primary content to enhance the atmosphere or mood, while SFX are to boost the expressiveness of key moments in a short period of time. As such, SFX are much shorter and need to explicitly synchronize with the picture. 

Previous studies add SFX to videos either by video-to-SFX matching \cite{arandjelovic2017look, lin2023soundify, wilkins2023bridging} or by video-to-SFX generation \cite{ghose2020autofoley, huang2023make, comunita2024syncfusion}, both at a \emph{holistic} level. They thus lack the ability of adding SFX to a specific moment.
Meanwhile, previous studies on video highlight detection \cite{lei2021detecting,li2024unsupervised} or
video moment retrieval \cite{moon2023query,liu2024r} consider only moment localization, leaving moment to SFX matching untouched.

We go beyond video-to-SFX matching by proposing a new task termed \underline{V}ideo \underline{D}ecoration with \underline{SFX} (\textbf{VDSFX}). In particular in the context of E-commerce field, VDSFX is to automatically add one or multiple SFX to key moments in a given e-commerce  video for better engaging experience. The task is thus more practical yet more challenging. As exemplified in Fig. \ref{fig:intro}, simply combining a state-of-the-art video moment detection model \cite{liu2024r} and moment-to-SFX matching is insufficient. 
In order to determine where in the video shall SFX be added, we need to know what SFX are available at hand. Let us imagine a video showing a dog and a cow in separate moments, yet the SFX set has cow sound only. While both moments are key moments of animal presence, only the cow moment is valid subject to the SFX set. Hence, the availability of candidate SFX has to be considered for both moment detection and moment-to-SFX matching. Such a mechanism is naturally missing when the two subtasks are performed sequentially. The questions of where in the video to add SFX and which SFX to add have to be jointly answered.

Naturally, no public data exists for the new task. To fill out the gap, we contribute \textbf{SFX-Moment}, a dataset for VDSFX. With over 16k highly selected videos, nearly 40k key moments and 356 distinct SFX in total, SFX-Moment is the first dataset for VDSFX, to the best of our knowledge.

\begin{figure*}
  \centering
  \includegraphics[width=\textwidth]{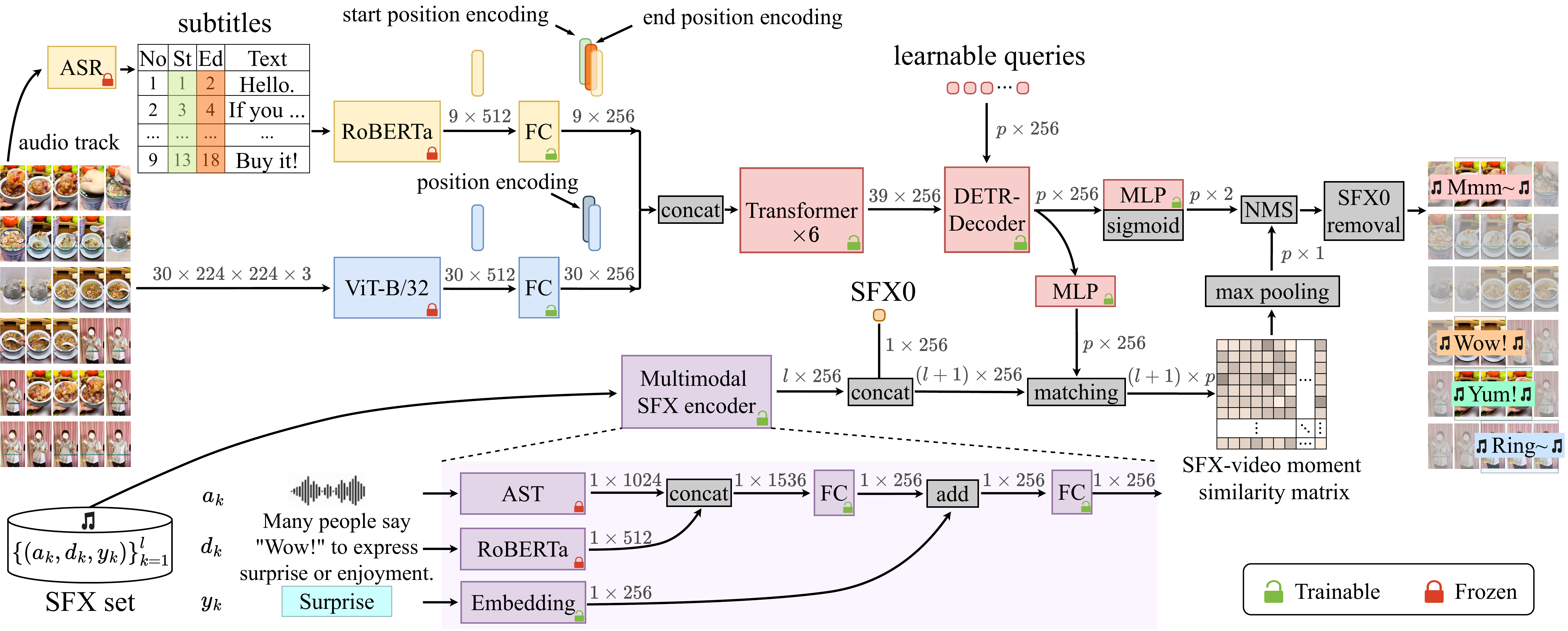}

  \caption{Diagram of our proposed D\&M method for VDSFX. The input video as an example consists of $30$ frames with $9$ subtitles. Each sound effect, indexed by $k$, is jointly represented by an audio clip $a_k$, a manually written short description $d_k$ and a categorical tag $y_k$.   SFX0 is a special token indicating ``no SFX". The ASR module and the visual / textual / audio backbones, \ie ViT / RoBERTa / AST, are all frozen. Non-trainable blocks are shown in gray. Best viewed on screen. } \label{fig:inference}
\end{figure*}

Inspired by the great success of DETR-based methods for video highlight detection ~\cite{lei2021detecting, moon2023query}, we propose \textbf{D\&M}, a DETR-based method that performs key moment \underline{d}etection and moment-to-SFX \underline{m}atching simultaneously, see Fig. \ref{fig:inference}. 
Despite its relatively simple network architecture, training D\&M is nontrivial due to the misalignment between the video and SFX representations and the inclusion of false negative examples by commonly used hard negative sampling. For more effective training, we propose moment-SFX matching (MSM) based pre-training for better initialization and tag-aware negative sampling (TaNS) to balance the hard and false negatives.
In sum, our major contributions are as follows:

Our key contribution can be summarized as follows:
\begin{itemize}
\item We introduce a new task, Video Decoration with SFX (VDSFX), aiming at detecting key moments in a given video and adding SFX for corresponding moments. A task-specific dataset SFX-Moment is developed. 
\item We propose a DETR-based method D\&M, which performs key moment detection and SFX matching simultaneously. Two task-specific training strategies are developed to ensure successful network training. 
\item Extensive experiments on SFX-Moment show the viability of the proposed method and its superior performance over multiple re-purposed baselines. 
\end{itemize}

\section{Related Work} \label{sec:related}

Tackling the VDSFX task requires answering the combined question of where in the video to add SFX and which SFX to add. While a complete answer remains to be developed, partial answers exist as video highlight detection / video moment retrieval to the first part of the question and video-to-SFX retrieval to the second part. We thus briefly review progresses in these topics.

\textbf{Video Highlight Detection} (VHD) is to detect interesting or salient moments in a given video. Earlier methods for VHD compute the salience score per video clip and then retrieve highly scored clips as highlights \cite{sun2014ranking,yao2016highlight, xiong2019less,badamdorj2021joint}. 
Viewing highlight moments as a specific object along the temporal dimension, \cite{lei2021detecting} extends DETR, originally developed for image object detection \cite{carion2020end}, to Moment-DETR for VHD and video moment retrieval (VMR). 
The success of Moment-DETR has inspired a number of follow-up studies \cite{xu2023mh, moon2023correlation, moon2023query, jang2023knowing,lin2023univtg, liu2024r}. 
Our work also gets inspriation from Moment-DETR, but goes one step further by adding a novel moment-to-SFX matching module, which is learned jointly with the key moment detection module. 

\textbf{Video-to-SFX Retrieval} aims to match SFX with a given video to create a more engaging multimedia experience. 
Depending on whether the video encoder and the SFX encoder are coupled, we categorize existing methods into two groups: one-tower \cite{arandjelovic2017look} and two-tower \cite{wilkins2023bridging}. 
\cite{arandjelovic2017look} introduces the $L^3$ Net, a one-tower model that uses a multi-modal feature fusion layer to determine the correspondence of a given video-SFX pair. \cite{cramer2019look} improves the $L^3$ net by exploiting various audio embeddings.
As for the two-tower methods, \cite{wilkins2023bridging} proposes to represent the audio input in a pre-trained CLIP image embedding space. \cite{lin2023soundify} assigns each sound effect with a label which is vectorized by the CLIP text encoder for representation learning. ImageBind \cite{girdhar2023imagebind} uses images / videos to bind other modalities, with SFX encoded by a pre-trained Audio Spectrogram Transformer (AST) \cite{gong2021ast}.
The above methods  perform video-to-SFX matching at a holistic level, lacking the capability to add SFX to specific moments.
The proposed D\&M method tackles the deficiency with a key moment detection head, effectively transforming the holistic video-to-SFX matching into a \emph{local} moment-to-SFX matching for the new VDSFX task.

\section{SFX-Moment: A Dataset for VDSFX}

To the best of our knowledge, no public dataset exists for VDSFX. To fill out the gap, we construct SFX-Moment, a set of 16,942 high-quality videos sampled from an E-commerce platform, see Table \ref{tab:dataset} . Each video has at least one key moment. For each moment, a meme sound effect is automatically extracted from the video creator's editing records. These records faithfully reflect a real user's thought about which part of a given video shall be viewed as key moments and which SFX shall be used to decorate these moments. Next, we detail the data construction process. 

\begin{table} [!htbp]
\renewcommand{\arraystretch}{1.1}
\centering
\setlength{\tabcolsep}{3pt}
\resizebox{\linewidth}{!}{
\begin{tabular}{@{}lrrrrr@{}}
\toprule
\textbf{Data split} & \textbf{Videos} & \textbf{Total duration} & \textbf{ASR subtitles} & \textbf{Key moments} & \textbf{Unique SFX}   \\
\midrule
{\fontsize{11}{1}\selectfont Total} & 16,942 & 215.8 hrs & 398,603 & 39,670 & 356  \\ [2pt]
{\fontsize{11}{1}\selectfont \textit{training}}    & 13,560 & 173.6 hrs & 320,508 & 31,836 & 356   \\
{\fontsize{11}{1}\selectfont \textit{validation}}   & 1,705  & 21.6 hrs & 39,929  & 4,021  & 291    \\
{\fontsize{11}{1}\selectfont \textit{testing}}         & 1,677  & 20.6 hrs & 38,166  & 3,813  & 283    \\
\bottomrule
\end{tabular}}
\caption{Statistics of our SFX-Moment dataset. The average video duration is 45.8s. The average number of SFX per video is 2.3. The average number of ASR subtitles per video is 23.5. We consider a closed set of SFX: a specific sound effect from the set can be applied to multiple videos.}
\label{tab:dataset}
\end{table}

\textbf{Data gathering}. We collected videos posted to an E-commerce video creation platform between Jan. 2023 and Aug. 2023, obtaining 100k videos initially. The videos have frame rate of 34 and image resolution of 1080$\times$ 1920. The mp3 audio files and metadata of the SFX including IDs and start-/end- times in the videos were also gathered, obtaining an initial set of 4k SFX. 

\textbf{Data cleaning}. We remove videos that fail to meet any of the following ad-hoc rules: 1) 200+ clicks; 2) 100+ follows; 3) duration between 3 seconds to 2 minutes; 4) With non-rare SFX (we consider a sound effect rare if it is used fewer than 10 times). 
Consequently, we preserve 16,942 videos, each with one or more SFX from a set of 356 unique SFX. 

\textbf{Data post-processing}. Per video, we use Whisper large-v3 \cite{radford2023robust}, a state-of-the-art model for automatic speech recognition (ASR), to transcribe the video narration into a sequence of timestamped sentences. 
Inspired by the successful use of music tags in video-to-music retrieval \cite{shin2017music, li2019query}, we manually categorize each sound effect to one of the following six tags: \emph{prompt}, \emph{transition}, \emph{humor}, \emph{action}, \emph{surprise} and \emph{others}, see Table~\ref{tab:tag}. Typical examples in the \emph{others} are ``Spooky Screaming'', ``Mosquito Flying'' and ``Mains Hum''. Furthermore,  per sound effect, a short description is manually written to reflect its  characteristics and potential use. Given the relatively small size of the SFX set, such manual annotation is affordable. Each sound effect, associated with an audio clip, a short description and a categorical tag, is multi-modal. 

\begin{table} [!thb] 
\renewcommand{\arraystretch}{1.1}
\centering
\setlength{\tabcolsep}{3pt}
\resizebox{\linewidth}{!}{
\begin{tabular}{@{}lrrrl@{}}
\toprule
\textbf{Tag} &  \textbf{SFX} & \textbf{Moments} & \textbf{Avg. duration} & \textbf{Tagging criterion} \\
\midrule
{\fontsize{12}{0}\selectfont \emph{prompt}}  &   77  & 17,574 & 2.2 s & {\fontsize{11}{0}\selectfont Attract attention when new things appear.}  \\
{\fontsize{12}{0}\selectfont \emph{transition}}  &    49 & 5,416 & 2.2 s  & {\fontsize{11}{0}\selectfont Transfer focus to the next scene.} \\
{\fontsize{12}{0}\selectfont \emph{humor}}   &    53 & 4,914 & 3.6 s  & {\fontsize{11}{0}\selectfont Enhance humor.} \\
{\fontsize{12}{0}\selectfont \emph{action}} &    71 & 4,225 & 2.3 s & {\fontsize{11}{0}\selectfont Dub for everyday activities.} \\
{\fontsize{12}{0}\selectfont \emph{surprise}} &  34 & 3,730 & 2.1 s  & {\fontsize{11}{0}\selectfont Express surprise or shock.}  \\
{\fontsize{12}{0}\selectfont \emph{others}}  &  72 & 3,811 & 2.8 s  & {\fontsize{11}{0}\selectfont Other SFX.}  \\
\bottomrule
\end{tabular}}
\caption{Tags for SFX categorization.}
\label{tab:tag}
\end{table}

\begin{figure}[!htbp]
    \centering
    \subfigure[] {\includegraphics[width=0.9\linewidth]{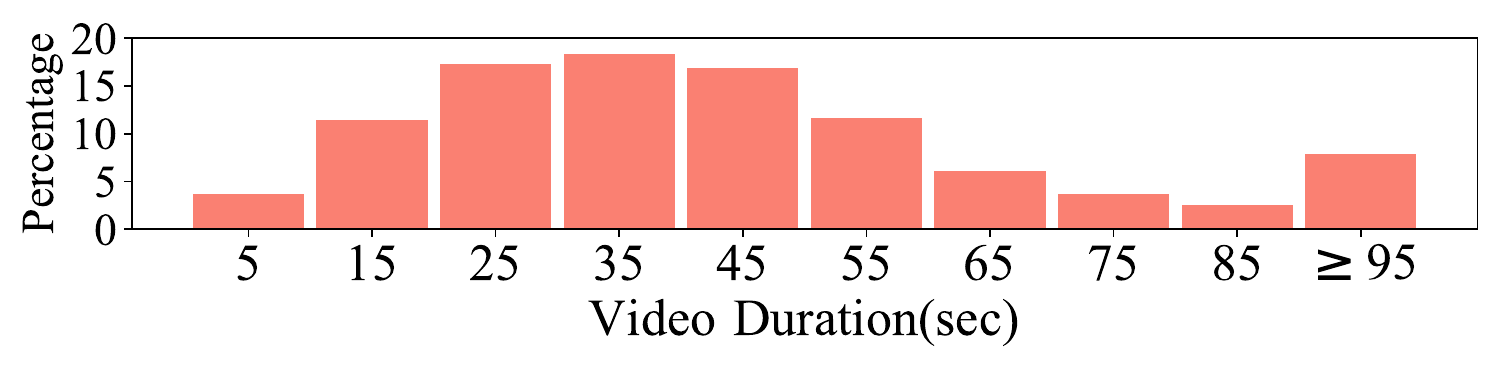}}
    \subfigure[] {\includegraphics[width=0.9\linewidth]{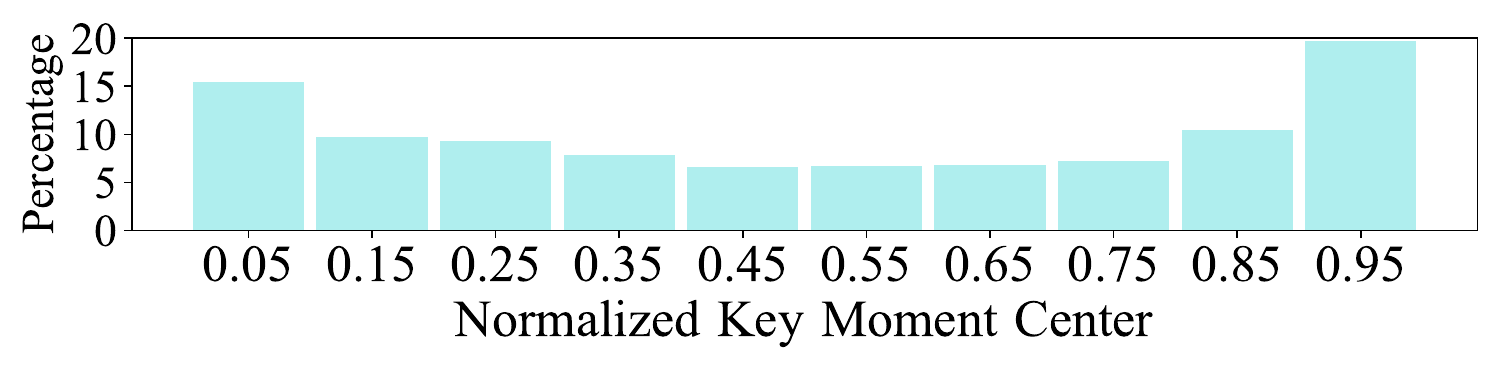}}
    \subfigure[] {\includegraphics[width=0.9\linewidth]{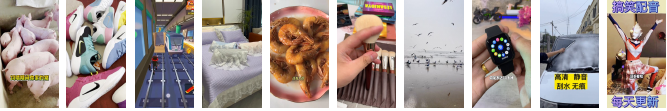}}
    \caption{
    Visualization of SFX-Moment. (a) Video duration. (b) Centers of key moments normalized by video length. (c) Snapshots of video samples. 
    }
    \label{fig:distribute}
\end{figure}

Note that the SFX-Moment videos are not significantly different from everyday videos, see Fig. \ref{fig:distribute}. The videos are categorized by 41 high-level tags, with \emph{agriculture}, \emph{fashion}, \emph{games}, \emph{home furnishing}, \emph{food}, \emph{beauty}, \emph{life}, \emph{digital}, \emph{cars}, \emph{funny} as the top-10 tags. A method working with such diverse video content needs to find patterns relevant and generalizable for VDSFX.

\section{Proposed D\&M Method} \label{sec:method}

A desirable method for VDSFX  shall find temporally \emph{where} in a given video to add a sound effect and \emph{what} sound effect to add. More formally, we use $v$ to denote the given video, with its visual content represented by $n$ evenly sampled frames $\{f_i\}_{i=1}^n$. An off-the-shelf ASR tool shall transcribe the associated narration into a sequence of $m$ timestamped sentences $\{(b_j, e_j, t_j)\}_{j=1}^m$, where $b_j$ and $e_j$ indicate the beginning and end timestamps, measured by frame indices, of sentence $t_j$. Suppose we have access to a set of $l$ SFX $\{(a_k, d_k, y_k)\}_{k=1}^l$, with $a_k$ indicating the audio clip of the $k$-th sound effect, $d_k$ as the short description and $y_k$ as the categorical tag, \emph{c.f.} Sec. 3. We get around the two challenges,  \ie key moment \underline{d}etection and moment-SFX \underline{m}atching, in a joint framework, dubbed D\&M. We describe the proposed network in Sec. \ref{ssec:inference}, followed by a task-specific training strategy in Sec. \ref{ssec:training}.

\footnotetext[4]{Chinese CLIP \cite{yang2022chinese}, as our data is in Chinese.}

\subsection{Network} \label{ssec:inference}

The network structure of D\&M is illustrated in Fig. \ref{fig:inference}. We largely follow the network structure of Moment-DETR~\cite{lei2021detecting}, using a multi-Transformer encoder for multi-modal feature extraction and a DETR-decoder for multi-head prediction. However, as Moment-DETR is targeted at fore-/back-ground moment classification, we need to introduce multiple VDSFX-specific modules as follows.

\subsubsection{Multi-modal video embedding}  Given the video and its subtitles as a multi-modal input, we first utilize a pre-trained CLIP model\footnotemark[4] to extract $n$ frame-level and $m$ sentence-level embeddings, respectively. As a rule of thumb, each frame / sentence is represented by its CLS token. To simplify our notation, we re-use $f$ and $t$ to denote the two sorts of embeddings. Each then goes through an FC layer for domain-adapted learning. The re-learned tokens,  with their positional embeddings $pe_0$ / $pe_1$ added, are concatenated to form an array of $n+m$ mixed-modal tokens denoted as  $\{\hat{f}_i\}_{i=1}^n$, $\{\hat{t}_j\}_{j=1}^m$. These tokens are fed to a multi-Transformer encoder to produce $n+m$ multi-modal tokens $\{z_i\}_{i=1}^{m+n}$. The above process can be expressed more formally as
\begin{equation} \label{eq:video-embedding}
\resizebox{.9\hsize}{!}{$
\left\{
\begin{array}{lll}
\{f_i\}_{i=1}^n & \leftarrow &  \text{ViT-B/32}(\{f_i\}_{i=1}^n),\\
\{t_j\}_{j=1}^m & \leftarrow &  \text{RoBERTa}(\{t_j\}_{j=1}^m),\\
\{\hat{f}_i\}_{i=1}^n & \leftarrow & \text{FC}_{512\times 256}(\{f_i\}_{i=1}^n) + pe_0, \\
\{\hat{t}_j\}_{j=1}^m & \leftarrow & \text{FC}_{512\times 256}(\{t_j\}_{j=1}^m) + pe_1, \\
\{z_i\}_{i=1}^{n+m} & \leftarrow & \text{Transformers}(\{\hat{f}_i\}_{i=1}^n, \{\hat{t}_j\}_{j=1}^m). 
\end{array}
\right. $}
\end{equation}

\subsubsection{Multi-modal SFX embedding}
Given a sound-effect triplet $(a, d, y)$, we first employ AST of ImageBind~\cite{girdhar2023imagebind}, the previously used CLIP and an embedding layer to extract the audio, sentence and tag features, indicated by $a$, $d$ and $e(y)$, respectively.  A multi-modal embedding of the sound effect, denoted as $s$, is obtained by a simple network as follows
\begin{equation} \label{eq:sfx-embedding}
\resizebox{.9\hsize}{!}{$
\left\{
\begin{array}{lll}
a & \leftarrow & \text{AST}(a),\\
d &  \leftarrow & \text{RoBERTa}(d), \\
e(y) & \leftarrow & \text{Embedding}_{l \times 256}(y),\\
s   & \leftarrow & \text{FC}_{\cdot \times 256}(\text{FC}_{\cdot\times 256}([a, d]) + e(y)).
\end{array}
\right. $}
\end{equation}
We thus obtain the SFX embeddings as $\{s_k\}_{k=1}^l$. 
Note that in addition to the normal SFX, we specify a special SFX0, with its trainable embedding $s_0$, to handle moments that cannot be matched with any of the normal SFX. 

\subsubsection{Key moment detection and embedding}
With the video and SFX embeddings available, we are now ready to invoke a DETR-decoder for key moment detection and embedding. In particular, the decoder uses an array of $p$ learnable queries $\{q_j\}_{j=1}^p$ to progressively conduct cross-attention based interaction with the video embeddings, resulting in $p$ output embeddings $\{\hat{q}_j\}_{j=1}^p$. The output is fed in parallel into two distinct MLPs, one for moment localization and the other for moment embedding. More formally, we have
\begin{equation} \label{eq:decoder}
\resizebox{0.9\hsize}{!}{$
\left\{
\begin{array}{lll}
\{\hat{q}_j\}_{j=1}^p & \leftarrow & \text{DETR-Decoder}(\{q_j\}_{j=1}^p, \{z_i\}_{i=1}^{n+m}), \\
\{pos_j\}_{j=1}^p & \leftarrow & \text{MLP}_{loc}(\{\hat{q}_j\}_{j=1}^p), \\
\{e_j\}_{j=1}^p & \leftarrow & \text{MLP}_{emb}(\{\hat{q}_j\}_{j=1}^p), 
\end{array}
\right. $}
\end{equation}
where $pos_j$ indicates (normalized) start-/end-positions of the moment predicted \wrt the $j$-th query and $e_j$ is a 256-d embedding of the predicted moment.

\subsubsection{Moment-SFX matching}
Given the $l+1$ SFX embeddings and $p$ moment embeddings, a pairwise cosine similarity calcluation yields an $(l+1)\times p$ similarity matrix $M$, see Fig. \ref{fig:inference}. Column-wise max pooling over $M$ allows us to obtain the best matched sound effect per moment. Non-Maximum Suppression (NMS), with an IoU threshold of 0.3, is performed to discard lower-confidence moments. Moments matched with SFX0 are also removed. The remained moment-SFX pairs are the final output.     

\subsection{Task-Specific Training} \label{ssec:training}

Although the network structure of D\&M is relatively simple, its training is nontrivial. Our early experiments show that training D\&M from scratch (with the pre-trained visual / textual / audio backbones frozen) results in a suboptimal model. Our conjecture is that the SFX and the video embeddings lack semantic alignment, resulting in poor matching during training, which in turn affects the DETR-decoder and the multi-Transformer encoder. To resolve the issue, we propose moment-SFX matching (MSM) based pre-training for better initialization of the SFX encoder and the video encoder. Moreover, in order to strike a proper balance between hard and false negative samples, we propose tag-aware negative sampling (TaNS) for more efficient training. 

\subsubsection{Pre-training for Better Initialization}

Given a key moment with $b$ and $e$ as its start-/end-frame index, we obtain its embedding $u$  by mean pooling over the index-bounded output of the multi-Transformer encoder, \ie $\text{mean-pooling}(\{z_i\}_{i=b}^e)$. Let $s$ be the embedding of the moment's corresponding sound effect, we write a mini batch consisting of $B$ moment-SFX pairs as $\{(u_i,s_i)\}_{i=1}^B$. For the special embedding $s_0$, we obtain its moment embedding as mean pooling over $\{z\}$ not covered by any SFX. 
MSM based pre-training can be achieved by minimizing an InfoNCE loss on the batch. Except for DETR-Decoder and two MLPs, the MSM based pre-training will give a head start for the subsequent end-to-end training of the D\&M network.

\subsubsection{Tag-aware Negative Sampling}

For constructing a mini batch for a given moment  and sound effect pair $(u,s)$, hard negative sampling is to select from the SFX set the top similar SFX other than the ground truth. We empirically find that such a sampling strategy tends to incorrectly include either false negatives or negatives too hard to learn from. To address the issue, we propose tag-aware negative sampling. The idea is simple as follows. For a candidate negative $s^-$, if its tag $y(s^-)$ is different from $y(s)$, the negative is reliable so its probability being sampled shall be proportional to its similarity to $u$, denoted as $sim(u,s^-)$. However, if $s^-$ shares the same tag as $s$, the risk of being false negative is relatively high. Therefore, its sampling probability will be \emph{inversely} proportional to its relative distance to the moment expressed as $|sim(u,s)-sim(u,s^-)|$. If $sim(u,s^-)$ is much larger than $sim(u,s)$, it is likely to be a false negative. By contrast, if  $sim(u,s^-)$ is much smaller, its value is much limited. Either case, the probability shall be reduced. More formally, we have
\begin{equation} \label{eq:tag-based}
\resizebox{.9\hsize}{!}{$
\text{p}(s^-) \propto \left\{
\begin{array}{cl}
 sim(u,s^-), & \text{if}~y(s^-) \neq y(s), \\
 -|sim(u,s)-sim(u,s^-)|, & \text{if}~y(s^-) = y(s).
\end{array}
\right. $}
\end{equation}

\subsubsection{Loss for end-to-end training}

For key moment localization, we follow Moment-DETR, computing the following two localization losses, \ie the L1 loss and the GIoU loss, to jointly measure the misalignment between the predicted and the ground-truth moments. As for moment-SFX matching, we again compute the InfoNCE loss. The three losses are equally combined. 

\section{Experiments} \label{sec:eval}

\subsection{Experimental Setup}

\subsubsection{Dataset}
We split randomly the 16,942 videos of SFX-Moment into three disjoint subsets: 13,560 (80\%) for training, 1,705 ($\sim$10\%) for validation and 1,677 ($\sim$10\%) for testing, see Table \ref{tab:dataset}.

\subsubsection{Evaluation Criteria}

The evaluation of VDSFX should consider both key moment detection and corresponding SFX matching simultaneously. We employ mean Average Precision (mAP) with a predefined IoU threshold (0.5 and 0.75 in this work and by default 0.5 unless otherwise stated).In particular, we compute ${\text{AP}}_{\text{SFX}}$, which measures the ranking of predicted key moments \wrt a given sound effect. We obtain ${\text{mAP}}_{\text{SFX}}$ by averaging over all the test SFX.  In a similar manner, we compute ${\text{AP}}_{\text{vid}}$, which measures the ranking of the matched and localized SFX \wrt a given test video, and ${\text{mAP}}_{\text{vid}}$ by averaging over all the test videos.

Moreover, we separately evaluate the capability of key moment detection. In addition to ${\text{AP}}_{\text{vid}}$ measuring the ranking of predicted key moments \wrt a given test video, we evaluate ${\text{AP}}_{\text{key}}$ which measures the ranking of predicted key moments \wrt a given moment class. In particular, we random choose 1k key moments in the test set and manually categorize each moment to one of the following four classes that occur frequently: \emph{introduction}, \emph{transition}, \emph{humour} and \emph{selling}, see Table \ref{tab:moment-classes}. We obtain ${\text{mAP}}_{\text{key}}$ by averaging over the four key moment classes.

\begin{table} [!htbp]
\centering
\resizebox{\linewidth}{!}{
\begin{tabular}{@{}lrl@{}}
\toprule
\textbf{Moment class} &  \textbf{Moments} & \textbf{Tagging criterion} \\
\midrule
\emph{introduction}     &    532      & Introduce products or their functions.  \\
\emph{transition}  &    134           & The current scene transfers to the next one.  \\
\emph{humour}   &    161              & The scene or the narration is humorous .  \\
\emph{selling} &    173               & Appeal for buying.  \\
\bottomrule
\end{tabular}}
\caption{Classes for key moment categorization.}
\label{tab:moment-classes}
\end{table}

\subsubsection{Implementation Details}
All experiments were conducted using PyTorch on eight NVIDIA TESLA V100 GPUs. The size of the video / moment / SFX embeddings is 256. The max length of frames and subtitles is empirically set to 12 and 40, respectively. The number of the learnable queries in DETR-Decoder is 10. 
 We use AdamW with an initial learning rate of 3e-4 with a weight decay  of 1e-4.
 The learning rate is decayed in a cosine schedule strategy. 
 The MSM sub-network for pre-training is trained  with a batch size of 512, while  D\&M is trained with a batch size of 256.

 \begin{table} [!htbp]
\centering
\resizebox{\linewidth}{!}{
\begin{tabular}{lrr|rr|rr}
\toprule
\multirow{2}*{\textbf{Method}} & \multicolumn{2}{c}{${\text{mAP}}_{\text{SFX}}$} & \multicolumn{2}{c}{${\text{mAP}}_{\text{vid}}$} & \multicolumn{2}{c}{Mean} \\
                            \cmidrule(r){2-3} \cmidrule(r){4-5} \cmidrule(r){6-7}
                      & @0.5 & @0.75 & @0.5 & @0.75 & @0.5 & @0.75 \\

\midrule
Sliding-Win                         & 9.3 & 3.4 & 31.5 & 14.6 & 20.4 & 9.0\\
$\text{Moment-DETR}+$               & 3.2 & 0.7 & 13.0 & 4.7  & 8.1 & 2.7\\
$\text{QD-DETR}+$                   & 2.7 & 0.9 & 10.2 & 4.0  & 6.4 & 2.4\\
$\text{UniVTG}+$                    & 9.4 & 4.2 & 31.7 & 17.4 & 20.6 & 10.8\\
$\text{R}^2\text{-Tuning}+$         & 9.6 & 3.7 & 32.3 & 17.7 & 21.0 & 10.7\\
\midrule
D\&M                                & \textbf{12.8} & \textbf{6.8} & \textbf{34.3} & \textbf{21.1} & \textbf{23.6} & \textbf{14.0} \\
\bottomrule
\end{tabular}}
\caption{Performance of different methods for VDSFX.}
\label{tab:baseline}
\end{table}

\begin{table}
\centering
\resizebox{\linewidth}{!}{
\begin{tabular}{lrr|rr|rr}
\toprule
\multirow{2}*{\textbf{Method}} & \multicolumn{2}{c}{${\text{mAP}}_{\text{vid}}$} & \multicolumn{2}{c}{${\text{mAP}}_{\text{key}}$} & \multicolumn{2}{c}{Mean}\\
                            \cmidrule(r){2-3} \cmidrule(r){4-5} \cmidrule(r){6-7}
                      & @0.5 & @0.75 & @0.5 & @0.75 & @0.5 & @0.75\\

\midrule
Sliding-Win                        & 43.7 & 17.9 & 23.5 & 4.4 & 33.6 & 11.2  \\
Moment-DETR                        & 31.6 & 12.0 & 16.4 & 3.1 & 24.0 & 7.6 \\
QD-DETR                            & 24.1 & 7.0  & 7.2  & 0.6 & 15.7 & 3.8 \\
UniVTG                             & 55.8 & 26.1 & 37.2 & 10.8 & 46.5 & 18.5 \\
$\text{R}^2$-Tuning                & 55.7 & 26.5 & 38.2 & 11.9 & 47.0 & 19.2 \\
\midrule
D\&M                                & \textbf{59.9} & \textbf{28.9} & \textbf{40.0} & \textbf{14.6} & \textbf{50.0} & \textbf{21.8} \\
\bottomrule
\end{tabular}}
\caption{Performance of different methods for Key Moment Detection.}
\label{tab:vhd}
\end{table}

\subsection{Comparison with Baselines}

\subsubsection{Baselines}

As aforementioned, no baseline method exists for VDSFX. A relatively straightforward solution is to first produce a number of candidate moments by a sliding window strategy,  and then use the trained moment-to-SFX matching(MSM) sub-network to find the best matches moment-SFX pairs. We term this naive baseline \texttt{Sliding-Win}. 

An obvious drawback of Sliding-Win is its extremely high computation cost. For more efficient key moment detection, we employ VHD methods to detect key moments before MSM. The  methods should be open-source and support video frames and subtitles as input, so we choose the following representative  methods that meets the requirements: 
\begin{itemize}
    \item 
Two DETR-based methods: Moment-DETR~\cite{lei2021detecting} and QD-DETR~\cite{moon2023query}, which can directly localize key moments. 
 \item  Two encoder-only methods: UniVTG~\cite{lin2023univtg} and $\text{R}^2$-Tuning~\cite{liu2024r} that predict boundary offsets of each frame and get final key moments through NMS. 
\end{itemize}

For a fair comparison, we train these methods on the same data as used by D\&M. Combining each of them with our pre-trained MSM sub-network forms four competitive baselines for VDSFX, which we denote as $\text{Moment-DETR}$+, $\text{QD-DETR}$+, $\text{UniVTG}$+ and $\text{R}^2\text{-Tuning}$+, respectively.

\subsubsection{Results} 

The performance of the baselines and the proposed D\&M method for VDSFX is reported in Table~\ref{tab:baseline}. Among the baselines, the two DETR-based methods perform clearly worse than their Encoder-only counterparts. Sliding-Win, while simple, shows relatively descent performance when compared to the other baselines. 
D\&M outperforms all the baselines in terms of all metrics. 

The performance of the methods on the Key Moment Detection subtask is summarized in Table~\ref{tab:vhd}. D\&M again performs consistently better than the baselines under all metrics. The effectiveness of D\&M is thus verified.

\begin{table}
\centering
\resizebox{\linewidth}{!}{
\begin{tabular}{llrrl}
\toprule
\# & \textbf{Setup} & ${\text{mAP}}_{\text{SFX}}$ & ${\text{mAP}}_{\text{vid}}$ & Mean  \\
\midrule
\rowcolor{gray!40}
0 & Full                         & \textbf{12.8} & \textbf{34.3} & \textbf{23.6} \\
\midrule
\multicolumn{4}{l}{\textbf{Pre-training:}}  \\
1 & \textit{w/o} pre-training     & 7.6 & 27.6 & 17.6(\textcolor{red}{-6.0}$\downarrow$) \\
2 & MSM $\rightarrow$ SSP          & 7.8 & 27.4 & 17.6(\textcolor{red}{-6.0}$\downarrow$) \\
3 & MSM $\rightarrow$ SFX-TP         & 8.1 & 27.9 & 18.0(\textcolor{red}{-5.6}$\downarrow$) \\
\midrule
\multicolumn{4}{l}{\textbf{When to use MSM?}}  \\
4 & pre-train $\rightarrow$ train                     & 9.8  & 30.9 & 20.4(\textcolor{red}{-3.2}$\downarrow$) \\
5 & pre-train $\rightarrow$ pre-train, train           & 12.6 & 33.5 & 23.1(\textcolor{red}{-0.5}$\downarrow$) \\
\midrule
\multicolumn{4}{l}{\textbf{Negative Sampling:}}  \\
6 & TaNS $\rightarrow$ Uniform Sampling                & 10.2  & 32.4 & 21.3(\textcolor{red}{-2.3}$\downarrow$) \\
7 & TaNS $\rightarrow$ HardNS  & 11.4  & 32.7 & 22.1(\textcolor{red}{-1.5}$\downarrow$) \\
8 & TaNS $\rightarrow$ one-side TaNS   & 11.8  & 32.8 & 22.3(\textcolor{red}{-1.3}$\downarrow$) \\
\midrule
\multicolumn{4}{l}{\textbf{Input modalities for video encoder:}}  \\
9 & \textit{w/o} Subtitles              & 6.6  & 24.9 & 19.7(\textcolor{red}{-3.9}$\downarrow$) \\
10 & \textit{w/o} Frames                 & 9.1  & 30.3 & 15.8(\textcolor{red}{-7.8}$\downarrow$) \\  
\midrule
\multicolumn{4}{l}{\textbf{Input modalities for SFX encoder :}}  \\
11 & \textit{w/o} audio                & 8.3  & 26.9 & 17.6(\textcolor{red}{-6.0}$\downarrow$) \\
12 & \textit{w/o} description                 & 10.3 & 30.1 & 20.2(\textcolor{red}{-3.4}$\downarrow$) \\  
13 & \textit{w/o} tag               & 10.6 & 30.7 & 20.7(\textcolor{red}{-2.9}$\downarrow$) \\  
\bottomrule
\end{tabular}}
\caption{Ablation studies. The mAPs are computed given IoU threshold of 0.5.}
\label{tab:ablation}
\end{table}

\subsection{Understanding D\&M}
Recall our D\&M achieves superior performance through proposed moment-SFX matching (MSM) based pre-training and Tag-aware negative sampling (TaNS). We analyse the effectiveness and rationality of these two schemes with several ablation studies. We also analyse the necessity of the individual  modalities in the video encoder and the SFX encoder. We report ${\text{mAP}}_{\text{SFX}}$, ${\text{mAP}}_{\text{vid}}$, and their mean. 

\begin{figure*}[!htb]
  \centering
  \includegraphics[width=\linewidth]{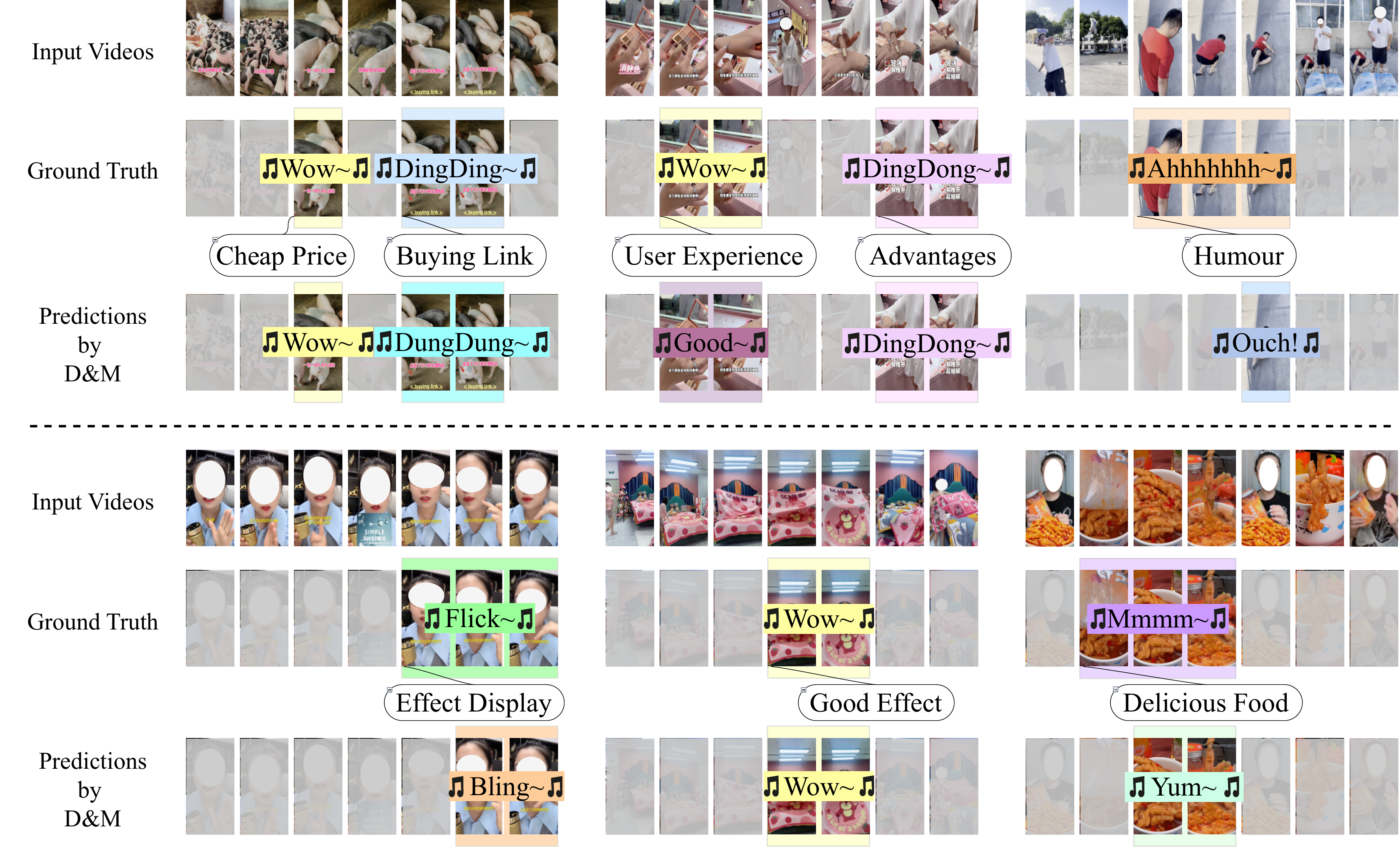}
  \caption{Some qualitative results. Best
viewed digitally.} \label{fig:vision}
\end{figure*}

\subsubsection{The Need of MSM based Pre-training} 
As Table \ref{tab:ablation} shows (Setup\#0 \emph{vs}. \#Setup\#1), no pre-training results in an absolute loss of 6.0 points in the overall performance. 

\subsubsection{The Need of TaNS}
Substituting unfiorm sampling for TaNS, indicated by Setup\#6, causes an absolute loss of 2.3 points. Using hard negative sampling (HardNS, Setup\#7) is better than uniform sampling, but remains worse than TagNS. Moreover, we also try one-side TaNS (Setup\#8), allowing only negatives that satisfy $y(s^-) \neq y(s)$. Its marginally better performance than HardNS (22.3 \emph{vs}. 22.1) suggests that excluding negatives that having the same tag as the positive SFX may help reduce false negatives. Still, TaNS is the best option.  

\subsubsection{MSM based Pre-training \emph{vs}. TaNS}
While both MSM based pre-training and TaNS are found to be necessary, the former contributes more to the good performance of D\&M.

\subsubsection{Alternatives to MSM for Pre-training?} 
We try two alternatives: \\
$\bullet$ Salience Score Prediction (SSP). SSP is previously used by \cite{lei2021detecting} as an auxiliary task for key moment judgement. In the current context, training SSP allows us to initialize the video encoder. \\
$\bullet$ SFX Tag Prediction (SFX-TP). The moment embedding $u$ is fed to an MLP to predict the SFX tag of the corresponding sound effect. Such a task let the video encoder be SFX-tag aware to some extent. 
The clearly lower performance of SSP (Setup\#2)  and SFX-TP (Setup\#3) suggests that initializing the video encoder alone is insufficient. 

\subsubsection{When to use MSM?} 
We try two options: \\
$\bullet$ Option-1: Using MSM directly as an auxiliary task for end-to-end training of D\&M (Setup\#4) \\
$\bullet$ Option-2: First using MSM for pre-training and re-using it as an auxiliary task for the subsequent end-to-end training (Setup\#5).

The better performance of Setup\#4 against Setup\#1 (20.4 \emph{vs}. 17.6) shows that using MSM as an auxiliary task is beneficial when training D\&M from scratch. Nevertheless, its lower performance than Setup\#0  (20.4 \emph{vs} 23.6) shows that MSM is best used for pre-training. The double use of MSM as Option-2 is unnecessary. 

\subsubsection{The influence of the input modalities for video embedding}
The clearly lower performance of Setup\#9 and Setup\#10 shows that both frames and subtitles matter, yet the former is more important. 

\subsubsection{The influence of the input modalities for SFX embedding}
Similar to the previous experiment, we observe that all the three modalities, \ie audio, description and tag, are necessary for effective SFX embeddings. Meanwhile, the audio modality is the most important, followed by the description modality and the tag modality. Such a rank is consistent with the (hidden) information each modality carries about SFX. 

Also, the ablation studies about the input modalities reveal the visual-audio based nature of the VDSFX task.

\subsubsection{Qualitative results}

Some qualitative results are shown in Fig. \ref{fig:vision}. For key moment detection, our D\&M's predictions are largely consistent with ground truth. Yet, the question of which SFX to choose remains challenging, leaving much room for future improvement.

\subsubsection{Efficiency analysis}

Given 20 frames and 40 subtitles per video, D\&M runs at a throughput rate of 27$\sim$34 videos per second, see Table \ref{tab:efficieny}.

\begin{table} [!htbp]
\centering
\resizebox{\linewidth}{!}{
\begin{tabular}{lrrrrr}
\toprule
\textbf{GPU} &  \textbf{Batch size} & \textbf{Embedding} & \textbf{Decoding} & \textbf{Overall} & \textbf{Videos per sec.} \\
\midrule
\text{V100} & 150 & 3.34s & 1.12s & 4.46s & 33.6 \\
\text{A10} & 100 & 1.88s& 1.06s & 2.94s & 34.0 \\
\text{2080 Ti} & 50 & 1.18s & 0.67s & 1.85s & 27.0 \\
\bottomrule
\end{tabular}}
\caption{Inference efficiency.}
\label{tab:efficieny}
\end{table}

\section{Concluding Remarks} \label{sec:conc}

We introduce in this paper a new task of video decoration with sound effects (VDSFX) and accordingly develop a task-specific dataset SFX-Moment. To tackle the new task, we propose D\&M, an end-to-end network that performs key moment detection and moment-to-SFX matching simultaneously. Extensive experiments on the SFX-Moment dataset allow us to draw conclusions as follows. Given the challenging nature of the task, moment-SFX matching (MSM) based pre-training and tag-aware negative sampling (TaNS) are important for effective training of the D\&M network. The proposed method compares favorably against multiple competitive baselines. With the developed dataset and method, our work opens up new possibilities for intelligent E-commerce video analysis and editing. 

\textbf{Limitations of this study}. Currently, each frame is represented by its CLS token. Such a holistic feature might lack the ability to effectively capture small-scale regions of interest. Consequently, for the adaption of D\&M to scenarios wherein fine-grained visual details matter, e.g., egocentric or third-person-view instructional videos, the current multi-modal video embedding module has to be improved, say by exploiting patch-level embeddings. Moreover, D\&M does not support interactive VDSFX, an important function for its real-world usability. How to enable the model to follow real-time user guidance deserves further investigation.

\section*{Acknowledgments}
This work was funded by NSFC (No. 62172420), Kuaishou and fund for building world-class universities (disciplines) of Renmin University of China.

\bibliography{my_work}

\end{document}